\documentclass{article}
\usepackage{spconf,amsmath,graphicx}
\newcommand{\etal}{\emph{et~al.}}
\usepackage{multirow}
\usepackage{color}
\usepackage{float}

\title{Transducer Adaptive Ultrasound Volume Reconstruction}
\name{Hengtao Guo\textsuperscript{1}, Sheng Xu\textsuperscript{2}, Bradford~J.~Wood\textsuperscript{2}, Pingkun~Yan\textsuperscript{1*}\thanks{* indicates corresponding author.}}
\address{\textsuperscript{1}Department of Biomedical Engineering and Center for Biotechnology and \\Interdisciplinary Studies, Rensselaer Polytechnic Institute, Troy, NY 12180, USA\\
\textsuperscript{2}National Institutes of Health, Center for Interventional Oncology, \\Radiology \& Imaging Sciences, Bethesda, MD 20892, USA}

\begin{document}
%
\maketitle
\begin{abstract}
Reconstructed 3D ultrasound volume provides more context information compared to a sequence of 2D scanning frames, which is desirable for various clinical applications such as ultrasound-guided prostate biopsy. Nevertheless, 3D volume reconstruction from freehand 2D scans is a very challenging problem, especially without the use of external tracking devices. Recent deep learning based methods demonstrate the potential of directly estimating inter-frame motion between consecutive ultrasound frames. However, such algorithms are specific to particular transducers and scanning trajectories associated with the training data, which may not be generalized to other image acquisition settings. In this paper, we tackle the data acquisition difference as a domain shift problem and propose a novel domain adaptation strategy to adapt deep learning algorithms to data acquired with different transducers. Specifically, feature extractors that generate transducer-invariant features from different datasets are trained by minimizing the discrepancy between deep features of paired samples in a latent space. Our results show that the proposed domain adaptation method can successfully align different feature distributions while preserving the transducer-specific information for universal freehand ultrasound volume reconstruction.
\end{abstract}
\section{Introduction}
\label{sec:intro}

\begin{figure}[htb]
\begin{minipage}[b]{1.0\linewidth}
  \centering
  \centerline{\includegraphics[width=0.85\textwidth]{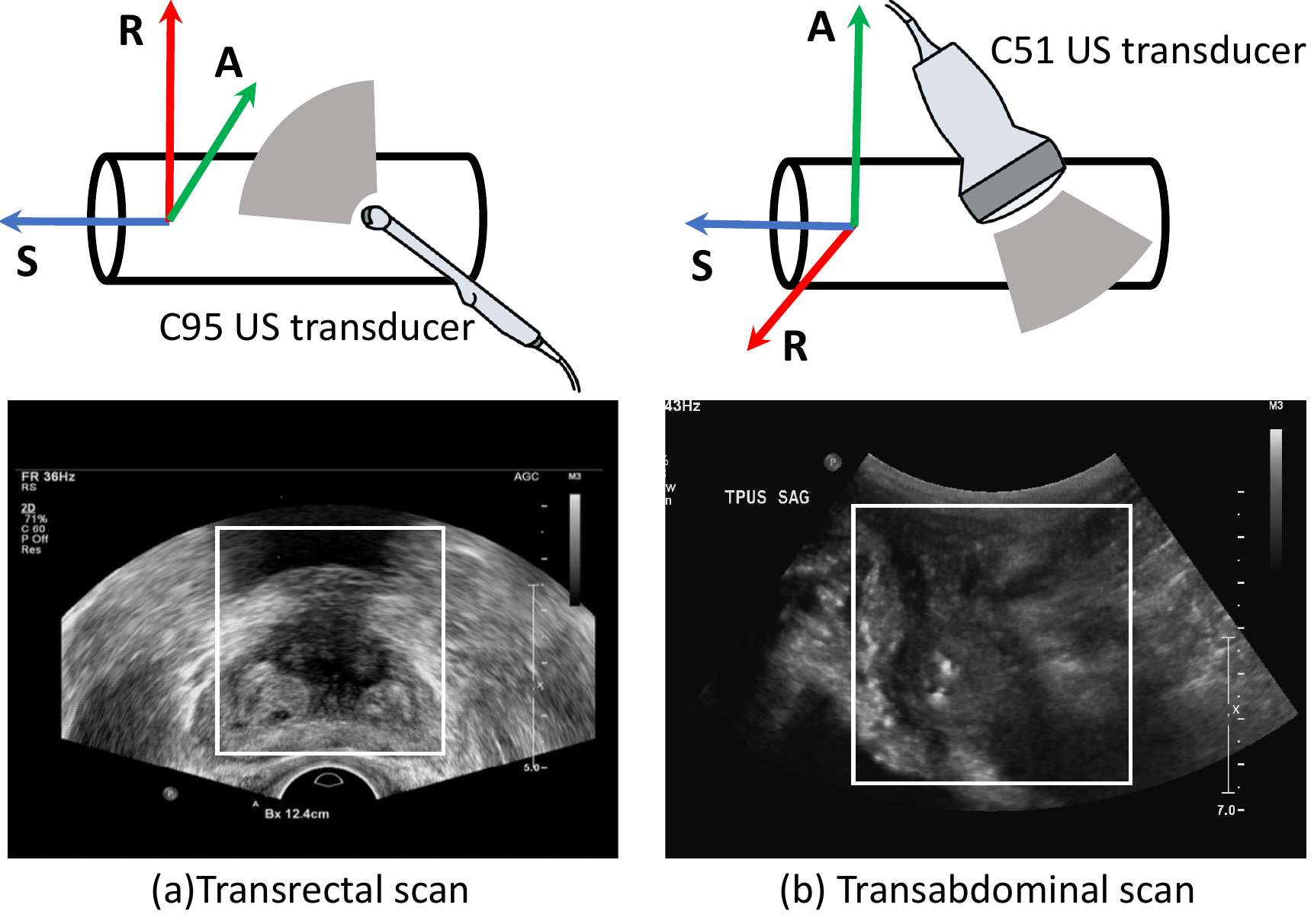}}
\end{minipage}
\caption{(a) Abdominal and (b) transrectal scans use different ultrasound transducers along different motion trajectories. The cylinder in the first row represents patient body, and ``RAS" indicates right, anterior and superior directions, respectively.}
\label{fig:scans}
\end{figure}

Ultrasound (US) is a commonly used medical imaging modality in various clinical applications. US possesses many advantages, such as low cost, portable setup, and the capability of real-time imaging. Compared to a sequence of 2D US frames, a reconstructed 3D US image volume can provide richer context information, which is often highly desired. Thus, efficiently reconstructing 3D US volume is a critical component in many interventional tasks, such as magnetic resonance imaging (MRI) and US fusion guided prostate biopsy~\cite{pinto2011magnetic,haskins2019learning,guo2020deep}. 

US volume registration from freehand ultrasound scans has traditionally implemented with tracking devices~\cite{wen2013accurate}, either an optical or electromagnetic (EM) tracking system, to record the position and orientation of US transducer in 3D space. Sensorless freehand scans takes a step further by removing the requirement of tracking devices. The original method was supported by the speckle decorrelation algorithms~\cite{tuthill1998automated}, which estimates elevational distance between neighboring US images based on the speckle patterns correlation. Recent advances of deep learning (DL) methods have shown superior performance in automatic feature extraction. Prevost \etal~\cite{prevost20183d} proposed to use convolutional neural network (CNN) to directly estimate the inter-frame motion between two 2D US frames for sensorless US volume reconstruction. In their latest work~\cite{wein2020three}, two DL-reconstructed volumes from transversal and sagittal views are co-registered for a better reconstruction result. Our recent work~\cite{guo2020sensorless} applies 3D CNN on a US video sub-sequence for better utilizing the temporal context information. 


Although CNNs have achieved promising results in US volume reconstruction, these methods suffer from severe performance degradation when applied to new datasets different from the training data. For example, as shown in Fig.~\ref{fig:scans}, both transrectal and transabdominal scans can be used to facilitate prostate cancer diagnosis, but they have distinct motion trajectories and imaging properties. One network trained on transrectal scans cannot produce satisfactory volume reconstruction on transabdominal scans. Here we define the source domain (transrectal scans) as the dataset  which serves as the training data of the CNN , and target domain (transabdominal scans) denotes the new dataset where the model is going to be applied to. Specifically, the domain shift is caused by difference between two datasets, leading to the model's decreased performance. Our target is to efficiently transfer the model trained on source domain to the target domain given limited target labeled samples. Thus, we formulate reconstructing US volume different US transducers as a domain adaptation problem in this work. 


Deep domain adaptation methods can be generally divided into two categories: adversarial-based and discrepancy-based~\cite{csurka2017domain}. The former methods propose to train a domain discriminator in an adversarial manner to enforce the feature vectors from both source and target domains to follow the same distribution~\cite{tzeng2017adversarial, hoffman2018cycada}. However, fooling the discriminator by generating mixed feature distributions does not help in our application. Our task is to make CNN accurately predict the relative position between two US frames. Merging the source and target feature distributions together without any high-level constraints contributes little to the task-specific feature learning: the adversarial strategy only pushes the target feature distributions close to that of the source, but does not enhance any specific feature learning that helps to accurately regress the inter-frame motion. 

In the image registration field, Mahapatra and Ge~\cite{mahapatra2020training} applied unsupervised domain adaptation for mono-modal medical image registration. An autoencoder was trained to extract latent feature vectors which are used for generating registered images through another generative adversarial network. Of note, the autoencoder was trained on chest X-ray images but the entire framework can be applied to registering images from other modalities such as brain MR images and retinal images. Another work by Zheng et al.~\cite{zheng2018pairwise} proposed a pairwise domain adaptation module to adapt the model from synthetic data to clinical data. Their primary assumption is that: if one X-ray image and one digitally reconstructed radiograph image were rendered from the same projection angle, the domain invariant feature extractor should extract consistent features from this real-synthetic image pair. In our work, on top of the discrepancy-based adaptation methods~\cite{saito2018maximum, tzeng2014deep}, we propose a novel paired-sampling strategy and use a discrepancy loss to transfer task-specific feature learning from source domain to target domain. We hypothesize that if two US video sub-sequences acquired using different transducers have similar motion trajectories, they should be close to each other in the latent feature space. Our contributions are summarized as follows:

\begin{enumerate}
\item 
We formulate our work on different US transducers as a domain adaptation problem. To the best of our knowledge, this reported work is the first to apply domain adaptation techniques to US volume reconstruction. 

\item
We propose a novel paired-sampling strategy with feature discrepancy minimization to facilitate model adaptation from the source to target domain. This strategy is specifically designed for registration-related domain adaptation problems.

\item
Our results demonstrate that the proposed method can extract domain-invariant features while preserving task-specific feature learning.
\end{enumerate}

\begin{figure}[t]
\begin{minipage}[b]{1.0\linewidth}
  \centering
  \centerline{\includegraphics[width=8.5cm]{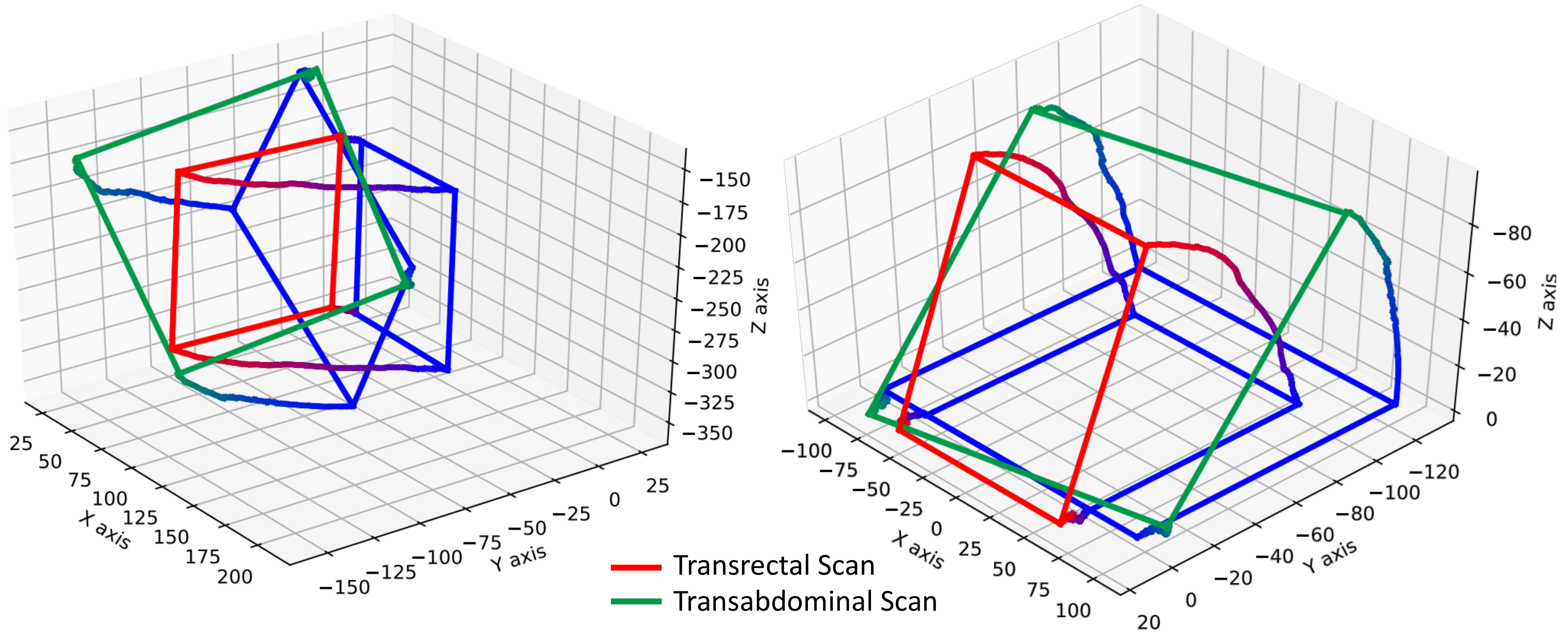}}
\end{minipage}
\caption{US video sequence trajectories in 3D before (left) and after (right) alignment. Blue frame indicates the first frame of a video sequence. Red and green label the last frames of a transrectal scan and a transabdominal scan, respectively.}
\label{fig:trajectory}
\end{figure}

\begin{figure}[t]
\begin{minipage}[b]{1.0\linewidth}
  \centering
  \centerline{\includegraphics[width=8.5cm]{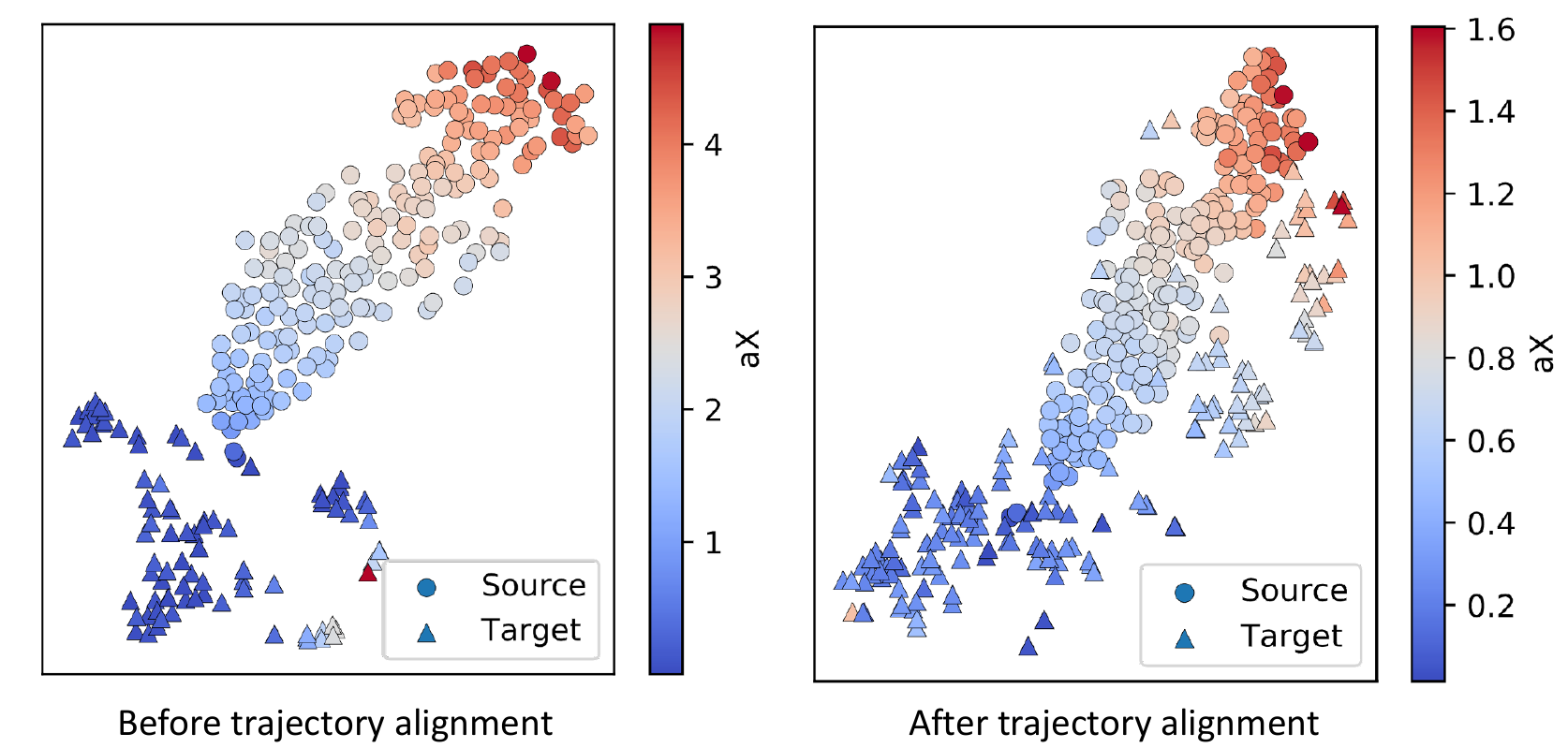}}
\end{minipage}
\caption{For each US video, we compute a mean DOF vector throughout the sequence and use t-SNE~\cite{maaten_tsne_2008} to project it into 2D space. The colorbar indicates the value of rotation $aX$ of each case, which is the most dominant motion direction. The trajectory alignment prevents the model's performance being influenced by the distribution gap in label space.}
\label{fig:doftsne}
\end{figure}

\begin{figure*}[tb]
    \centering
    \includegraphics[width=1\textwidth]{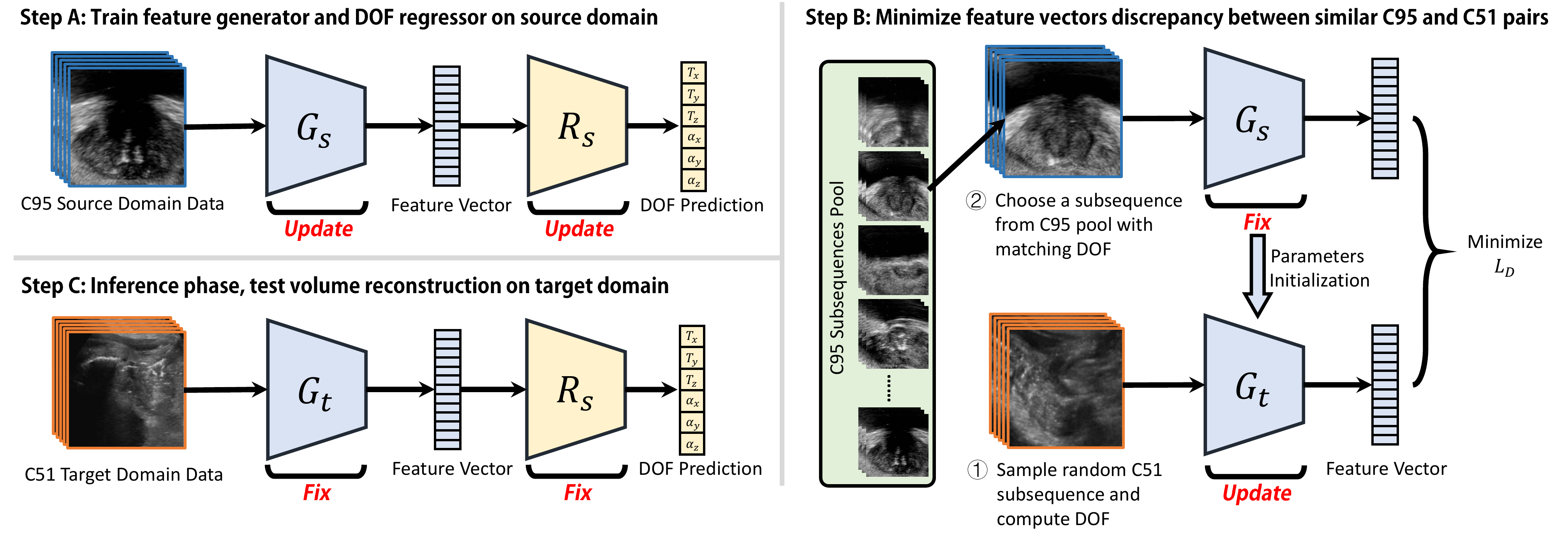}

\caption{Three steps of the proposed transducer adaptive ultrasound volume reconstruction (TAUVR) method, where transducer adaptation is achieved by minimizing the discrepancy between similar pairs sampled from both the source and target domains.
}
\label{fig:pipeline}
\end{figure*}

\section{MATERIALS AND METHODS}
\label{sec:methods}

\subsection{Transformation Space Alignment}

A primary task in 3D ultrasound reconstruction is to obtain the relative spatial position of two or more consecutive US frames. Consider a small subsequence containing $N$ consecutive frames as one sample unit, we can compute a relative transformation matrix and decompose it into 6 degrees of freedom (DOF) $Y={\left \{t_x,t_y,t_z,\alpha_x,\alpha_y,\alpha_z \right \}}$, which contains the translations in millimeters and rotations in degrees. The network takes one video subsequence as the input for estimating the transformation parameters. We use each subsequence's corresponding DOF vector as the groundtruth label~\cite{guo2020sensorless} during the training process. 

Since US transducers may have very different scanning trajectories for different applications (as in Fig.~\ref{fig:trajectory}), this large motion difference will create label bias and can substantially impair the network performance. To alleviate this problem, we add a pre-processing step to roughly align the video sequence trajectory in 3D space. More precisely, we first scale the US videos to the same resolution and align the first frame of the video sequences to the same position. The sequence rotating center (transducer's head) is overlapping with the origin $(0, 0, 0)$ of the 3D coordinate system. Thus, the label distributions of source domain and target domain are aligned together as in Fig.~\ref{fig:doftsne}. Before the trajectory alignment, the source and target DOF label distributions are separated into two clusters; after the alignment, the label distributions are merged together, showing a smooth $aX$ transition pattern. The trajectory alignment ensures that the model's performance will not be impaired by the gap in label distributions.

\subsection{Ultrasound Transducer Adaptation}

We denote our source domain dataset (transrectal scans) as $\{X_s|Y_s\}$, where each image sample $X_s$ represents a subsequence of $N=5$ consecutive frames and its correspond label $Y_s$ is a 6 DOF vector. In addition, we have another labeled but much smaller dataset on target domain (transabdominal scans) $\{X_t|Y_t\}$. Our proposed method for transducer adaptive ultrasound volume reconstruction (TAUVR) includes three consecutive steps as shown in Fig.~\ref{fig:pipeline}.

\textbf{Step A:} A convolutional feature extractor $G_s$ and a DOF regressor $R_s$ are trained in the source domain in an end-to-end fashion~\cite{guo2020sensorless}. The input to $G_s$ is a $N\times W\times H$ subsequence tensor and the output is a 2048D feature vector. The DOF regressor is a linear layer that outputs 6 values for DOF regression. $G_s$ and $R_s$ are jointly trained by minimizing the mean squared error (MSE) loss between network's output and groundtruth DOF labels.

\textbf{Step B:} In this step, we train a feature extractor $G_t$ on target domain which produces both domain-invariant feature while preserves task-specific information. $G_t$ is initialized with the parameters of $G_s$ and shares the identical structure, and $G_s$'s parameters are fixed in this step. 
\begin{table*}[htb]
    \begin{minipage}[b]{\linewidth}
    \centering
    \centerline{\includegraphics[width=\linewidth]{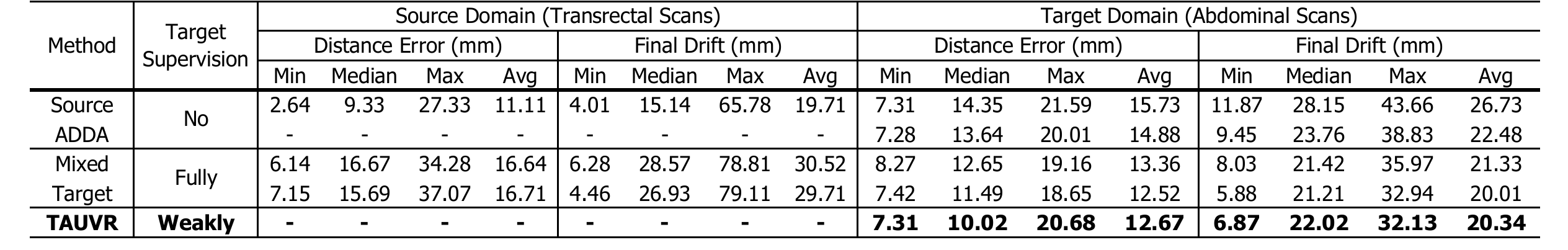}}
    \end{minipage}
    \caption{Perfomance of different methods on both source domain and target domain.}
    \label{tab:my_label}
\end{table*}
We first create a source domain subsequence pool in which every transrectal video subsequence has a corresponding DOF label vector. During adaptation training, for every random target subsequence sample $x_t$, we compute its DOF vector $y_t$ based on labeling information. Next, we search in the pool to find a source domain subsequence $x_s$ that has the closest motion vector as $y_t$. With this paired subsequence serve as the input to corresponding networks, we yield a pair of latent feature vectors denoted as:
\begin{equation}
    v_{s}=G_{s}\left ( x_{s} \right ), v_{t}=G_{t}\left ( x_{t} \right )
\label{eq:vec}
\end{equation}

$G_t$ is trained by minimizing the discrepancy loss $L_D$, which is the $L2$ norm between the two generators' output feature vectors:
\begin{equation}
    L_{D}=\frac{1}{P}\sum_{p=1}^{P}\left \| v_{s}^{p}-v_{t}^{p} \right \|_{2}
\label{eq:mmd}
\end{equation}
where $P$ denotes the total number of sampled pairs within one training epoch. The intuition of this paired sampling strategy is to establish correspondence between source and target subsequences: when two subsequences from different domains have similar motion, we expect their extracted feature vectors to be close to each other in the latent space. This paired-sampling strategy takes rich information in the labeled source dataset as a reference to guide task-specific features learning in the target domain. Since the labels of target domain data are only used for sampling subsequence pairs while do not directly contribute to the loss function, we categorized our strategy as a weakly-supervised method.

\textbf{Step C:} The final step is also the inference testing phase on target domain data and does not involve any parameters update. The networked used in this step is the concatenation of $G_{t}$ from Step B and $R_{s}$ from Step A. For a full-length US video sequence in the target domain test set, we use a sliding-window procedure to get the DOF motion vector prediction for every subsequence. By placing each frame into 3D space accordingly, a 3D US image volume can be reconstructed. The testing phase does not require any tracking devices and CNN estimates US frames relative position.


\section{Experiments}

\subsection{Settings}
All the data utilized in this study are acquired by the Nation Institute of Health (NIH) from IRB-approved clinical trial: 

\textbf{Source domain} contains 640 transrectal US video sequences, with each frame labeled a corresponding positioning matrix captured by EM-tracking device. An end-firing C95 transrectal ultrasound transducer captures axial images by steadily sweeping through the prostate from base to apex. The dataset is split into 500, 70 and 70 cases as training, validation and testing, respectively. 

\textbf{Target domain} contains 12 transabdominal US video sequences acquired by C51 US transducer. 9 cases are used for training in Step B and the network's parameters are saved after every epoch. 3 cases are used for testing in Step C.

Networks are trained for 300 epochs with batch size $K=24$ using Adam optimizer~\cite{kingma2014adam}. Each US frame is cropped without exceeding the imaging field (white bounding box in Fig.~\ref{fig:scans}) and then resized to $224 \times 224$. The entire pipeline is implemented using the publicly available PyTorch library~\cite{pytorch}. 


\subsection{Results and Discussions}

We present 4 baseline methods for comparison. As in Table~\ref{tab:my_label}, model on ``Source" was trained on source domain and then directly tested on target domain; ``Target" works in the opposite way; ``Mixed" is trained on merged source and target domain using all available label for supervision; ``ADDA"~\cite{tzeng2017adversarial} uses unsupervised adversarial domain adaptation method to extract domain-invariant features. The proposed TAUVR achieved significantly lower average distance error and final drift comparing to both ``Source" and ``ADDA". It is also comparable to the results of ``Target" while the latter still have a huge domain shift problem between source and target domain because of the model's overfitting to the transabdominal dataset.

\begin{figure}[h]
\begin{minipage}[b]{1.0\linewidth}
  \centering
  \centerline{\includegraphics[width=\textwidth]{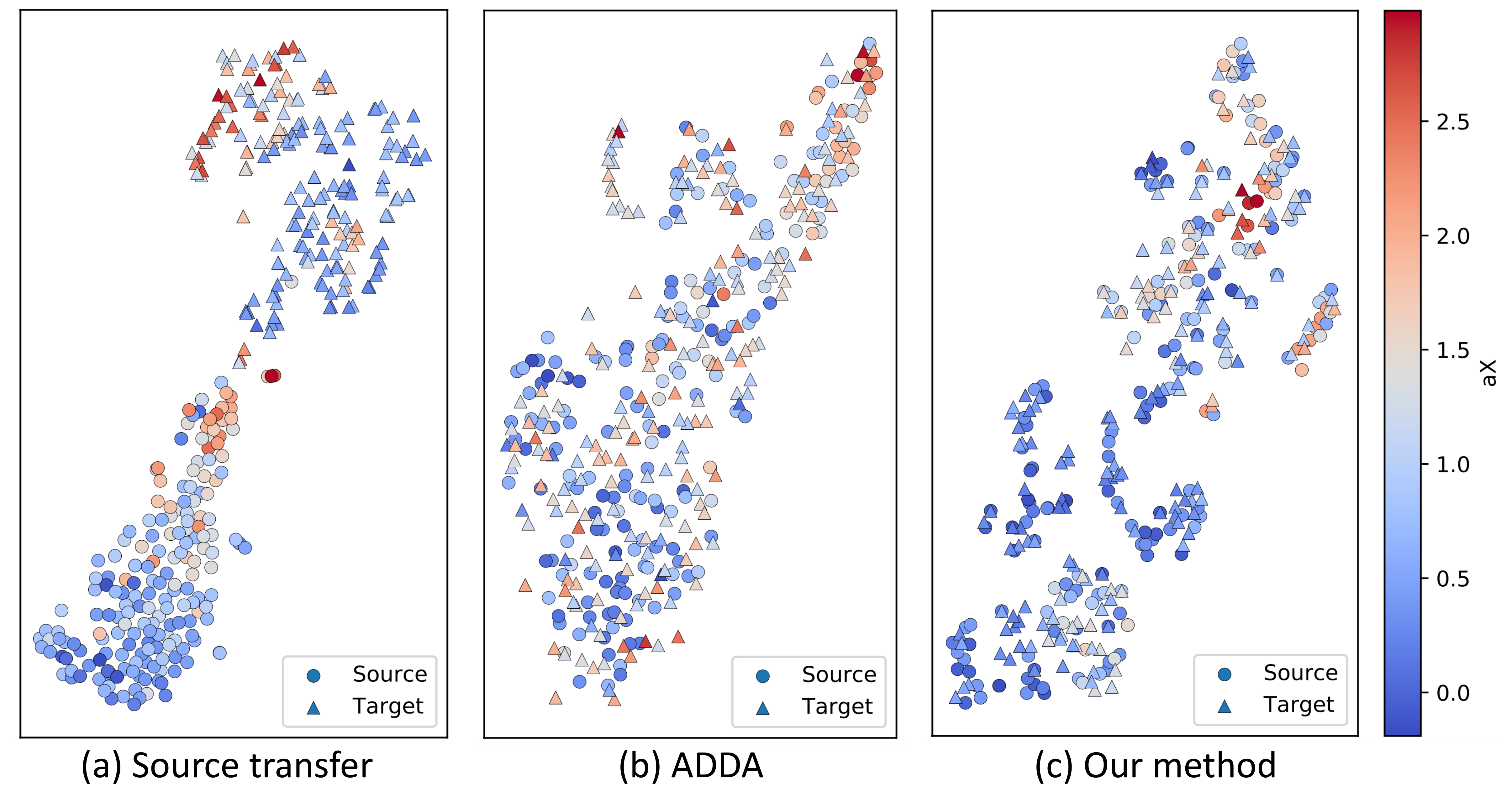}}
\end{minipage}
\caption{t-SNE projections of the latent feature vectors obtained using (a) source domain model only, (b) ADDA, and (c) our proposed method TAUVR.}
\label{fig:vectortsne}
\end{figure}

In addition, we evaluate the features quality through latent vector projections. As shown in Fig.~\ref{fig:vectortsne}, 2D tSNE projections of the extracted 2048D feature vectors are plotted, using the most dominant motion $aX$ for color encoding. 

On Fig.~\ref{fig:vectortsne}(a), points from source domain and target domain are roughly separated into two clusters, and within each cluster there exits a continuous changing pattern in $aX$ encoding. This indicates that (1) the network trained on source domain exhibit an obvious domain gap on target data, and (2) the network, however, still preserves the task-specific information in feature vectors.  

On Fig.~\ref{fig:vectortsne}(b), we observe that the distributions of two domains have been merged together through ADDA~\cite{tzeng2017adversarial}, as the adversarial training strategy tries to fool the domain discriminator by generating ``domain-invariant" features. However, since unsupervised learning poses no constraint on task-specific feature learning, the smooth color transition pattern disappears in the target domain (triangles), resulting in uninformative feature learning.

Our proposed method on Fig.~\ref{fig:vectortsne}(c) both merges the distributions of both domains and still keeps a gradual color transition in $aX$ for each domain. These phenomenons suggest that being benefited by the paired-sampling strategy, the network is extracting domain-invariant features while still preserves task-specific feature learning in both domains.

\begin{figure}[h]
\begin{minipage}[b]{1.0\linewidth}
  \centering
  \centerline{\includegraphics[width=\textwidth]{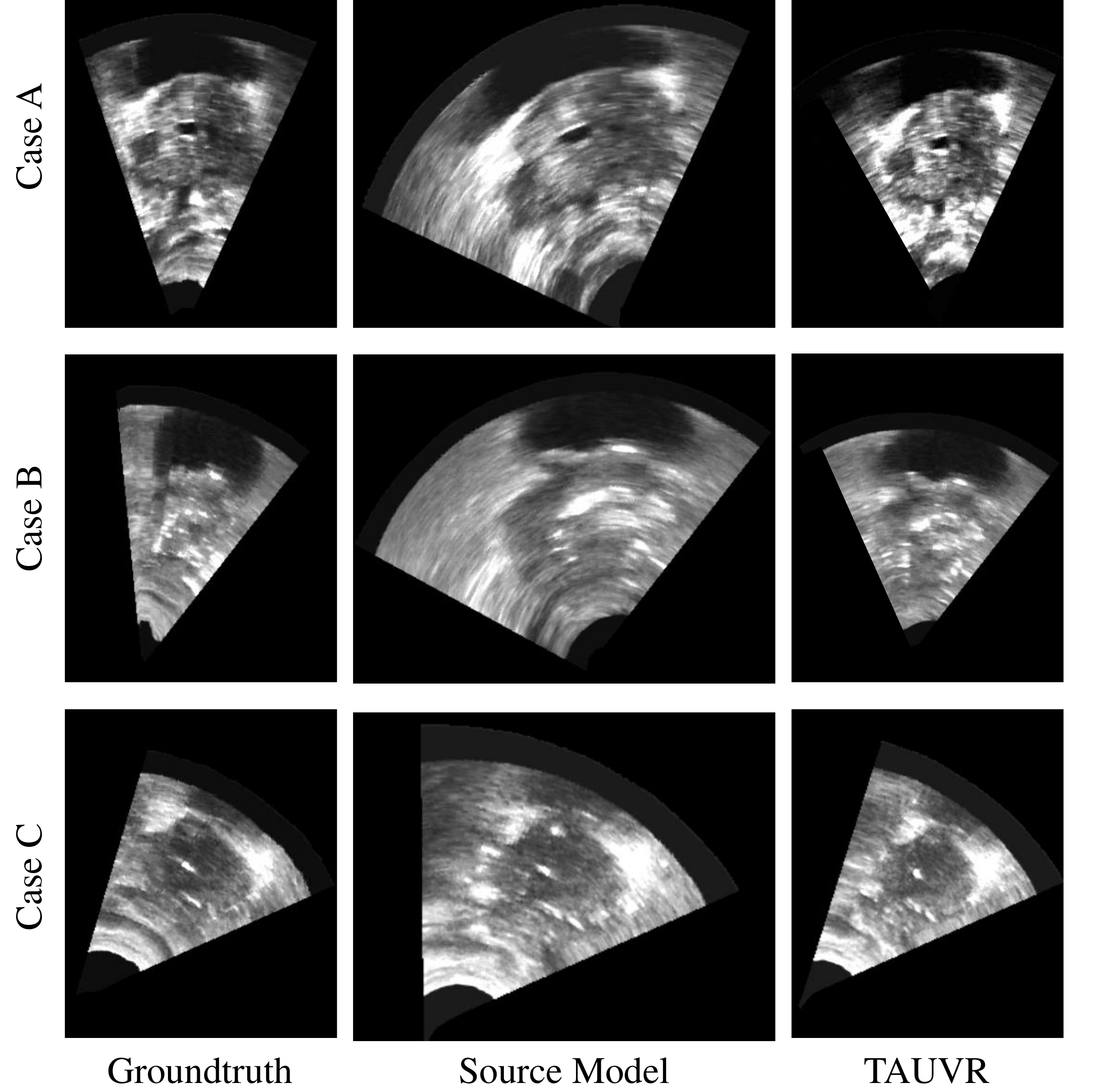}}
\end{minipage}
\caption{Sagittal view of the reconstructed ultrasound volumes from 3 testing cases in the target domain (transabdominal scans). The model trained on source domain produces very deviated reconstruction result in the target domain (middle column). By applying the proposed TAUVR (right column), the reconstruction is much closer to the groundtruth volume.}
\label{fig:volume}
\end{figure}

\subsection{Volume Reconstruction}

We present the sagittal view of the reconstructed volumes for quality assessment in Fig.~\ref{fig:volume}. All three test cases in the target domain (transabdominal scans) are presented by rows. From left to right, each column represents the reconstruction results from groundtruth labels, model trained only on source domain and our proposed TAUVR. As shown in the figure, by directly applying source model to the target data, the deep neural network may exhibit a over-fitting pattern that produce transducer trajectory prediction very close to that of the source domain. In other words, the trajectory prediction is deviated from the actual trajectory in the transabdominal scans. By incorporating our pairwise domain adaptation methods, the third column (TAUVR) produces visually much closer volume reconstruction comparing with the groundtruth.


\section{CONCLUSIONS}
In this paper, we presented a novel pair-sampling strategy to enhance task-specific feature learning in target domain, using matched source domain samples as reference. The proposed transducer adaptive method (TAUVR) allows sensorless ultrasound volume reconstruction, yielding a network that is capable of extracting domain-invariant features and preserve task-specific feature learning. The proposed method achieves promising results on target domain while the performance does not degrade on source domain. A more detailed evaluation of the proposed method for additional datasets will be provided in a comprehensive future work.

\label{sec:typestyle}




\section{Compliance with Ethical Standards}
\label{sec:compliance}
This research study was conducted retrospectively using human subject data acquired through an IRB-approved clinical study, in accordance with the ethical standards of the institutional and/or national research committee of where the studies were conducted.

\section{Acknowledgments}
\label{ssec:acknowledgments}
All authors have no conflict of interest to report. This work was partially supported by National Institute of Biomedical Imaging and Bioengineering (NIBIB) of the National Institutes of Health (NIH) under awards R21EB028001 and R01EB027898, and through an NIH Bench-to-Bedside award made possible by the National Cancer Institute.
\bibliographystyle{IEEEbib}
\bibliography{refs}

\end{document}